\def\year{2019}\relax
\documentclass[letterpaper]{article} 
\usepackage{aaai19}  
\usepackage{times}  
\usepackage{helvet}  
\usepackage{courier}  
\usepackage{url}  
\usepackage{graphicx}  
\usepackage{booktabs} 
\usepackage{tabularx}
\usepackage{arydshln}
\usepackage{wrapfig}
\usepackage{amsmath}
\usepackage{amsfonts}
\usepackage{graphicx}
\usepackage{subcaption}
\usepackage{algorithm}
\usepackage{algorithmic}
\usepackage{latexsym}
\usepackage{multirow}
\usepackage{tablefootnote}
\usepackage{caption}

\begin{document}
%
\title{DLGNet: A Transformer-based Model for Dialogue Response Generation}
\author{Oluwatobi O. Olabiyi and 
Erik T. Mueller\\
Capital One Conversation Research, Vienna, VA\\
\{oluwatobi.olabiyi, erik.mueller\}@capitalone.com
} 

\maketitle
\begin{abstract}
Neural dialogue models, despite their successes, still suffer from lack of relevance, diversity, and in many cases coherence in their generated responses. 
These issues can attributed to reasons including (1) short-range model architectures that capture limited temporal dependencies, (2) limitations of the maximum likelihood training objective, (3) the concave entropy profile of dialogue datasets resulting in short and generic responses, and (4) the out-of-vocabulary problem leading to generation of a large number of $<$UNK$>$ tokens. On the other hand, transformer-based models such as GPT-2 have demonstrated an excellent ability to capture long-range structures in language modeling tasks.  
In this paper, we present DLGNet, a transformer-based model for dialogue modeling. We specifically examine the use of DLGNet for multi-turn dialogue response generation. In our experiments, we evaluate DLGNet on the open-domain Movie Triples dataset and the closed-domain Ubuntu Dialogue dataset. DLGNet models, although trained with only the maximum likelihood objective, achieve significant improvements over state-of-the-art multi-turn dialogue models. They also produce best performance to date on the two datasets based on several metrics, including BLEU, ROUGE, and distinct n-gram. Our analysis shows that the performance improvement is mostly due to the combination of (1) the long-range transformer architecture with (2) the injection of random informative paddings. Other contributing factors include the joint modeling of dialogue context and response, and the 100\% tokenization coverage from the byte pair encoding (BPE).

\end{abstract}

\section{Introduction}
\label{introduction}
Recent successes of pretrained transformer-based language models, such as BERT \cite{Devlin2019}, GPT(-2) \cite{Radford2018,Radford2019}, Transformer-XL \cite{Dai2019}, XLNet \cite{Yang2019}, and ERNIE(2.0) \cite{Sun2019,Sun2019b}, have led to state-of-the-art performance on many natural language understanding (NLU) tasks including sentence classification, named entity recognition, sentence similarity, and question answering. The exceptional performance of transformer-based language models is due to their ability to capture long-term temporal dependencies in the input sequence. This attribute should be very beneficial to dialogue modeling, especially in multi-turn scenarios. Most of the existing neural dialogue response generation models are based on recurrent neural networks \cite{Sutskever2014,Vinyals2015,Li2016,Serban2016,Xing,Serban2017,Serban2017a,Li2016c,Zhang2018,Olabiyi2018,Olabiyi2019}.     

These models have yielded promising results by generating mostly coherent responses given the dialogue context. However, most of them, including the state-of-the-art models trained with naturalistic dialogue data, still perform well below the human level. Generated responses tend to be either generic, out-of-context, or disproportionately short. Previous work points to some causes of these limitations:

\begin{figure}[t]
\begin{center}
\centerline{\includegraphics[width=0.95\linewidth]{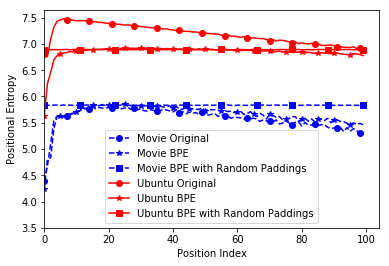}}
\caption{\textbf{Positional Entropy for Movie and Ubuntu datasets -}
Applying a greedy training objective to the original and BPE datasets can achieve low overall entropy just by overfitting to low entropy regions, resulting in short and generic responses. Injecting random paddings into the data does not suffer from this problem and can be used to train transformer architectures due to their lack of recurrent propagations. }
\label{entropy}
\end{center}
\vskip -0.3in
\end{figure}

\textit{i) Training data:} Human conversations contain a large number of generic, uninformative responses, giving rise to  word-level syntactic and utterance-level semantic redundancy. The syntactic redundancy is evident from a non-uniform sequence entropy profile, that is concave with respect to token position, with the tokens at the beginning and end of a sequence having lower entropy than those in the middle (see Fig. \ref{entropy}). This initial positive energy gradient can create learning barriers leading to a poor calibration of the model's output distribution, and is a major contributing factor to the short, generic outputs in existing dialogue models \cite{Olabiyi2019b}.

\textit{ii) Short-range Model Architecture}: Earlier conversation models are single-turn Seq2Seq architectures \cite{Sutskever2014,Vinyals2015,Li2016} that fails to capture long-term temporal dependencies across conversation turns. Such models tend to fail in multi-turn scenarios \cite{Li2016c}, generating repetitive responses that are dull and generic. The use of multi-turn Seq2Seq models, such as the hierarchical recurrent encoder decoder (HRED) architecture, tried to address this problem \cite{Serban2016,Xing,Serban2017,Serban2017a,Olabiyi2018,Olabiyi2018b,Olabiyi2019}. The recurrent architecture, however, due to the gradient vanishing problem with backpropagation through time, limits the maximum number of turns and the number of word tokens in each turn that are used during training.

\textit{iii) Out-of-vocabulary Problem:} One the major and often overlooked limitations of existing dialogue models is the limitations of the input/output representation \cite{Radford2019}. The data preprocessing used in existing dialogue models includes word-level tokenization and lowercasing with less frequent (usually more informative) words mapped to the out-of-vocabulary token $<$UNK$>$ and thus restrict the space of the input and output texts that can be modeled. This is especially problematic for closed-domain datasets with lots of technical jargon, where preprocessing yields a large number of $<$UNK$>$ tokens in both training and inference. Unfortunately, using character-level representations with 100\% coverage requires gradient backpropagation through a very long sequence, which is impractical for existing recurrent architectures.  

\textit{iv) Exposure Bias:} Similar to language and machine translation models, traditional conversation models are trained with the model input taken from the ground truth rather than a previous output (a method known as \textit{teacher forcing} \cite{Williams1989}). During inference, however, the model uses past outputs,~i.e., is used autoregressively. This is problematic in the dialogue setting since the learning task is unconstrained \cite{Lowe2015}. In particular, there are several suitable target responses per dialogue context and vice versa. This discrepancy between training and inference is known as \textit{exposure bias} \cite{Williams1989,Lamb2016} and significantly limits the informativeness of the responses as the decoding error compounds rapidly during inference.

\textit{v) Training Objective:} Existing dialogue models learn the conditional distribution of the response given the context (either single- or multi-turn), from the maximum likelihood estimation (MLE) \cite{Sutskever2014,Vinyals2015,Serban2016,Olabiyi2018}. Due to the redundant nature of dialogue data and the greedy nature of MLE, the model usually learns just a simple mapping between the context and response, which yields generic responses. Alternative training frameworks that complement MLE with other constraints, such as generative adversarial networks, reinforcement learning, and variational auto-encoders, focus on modifying the conditional response distribution to encourage diversity \cite{Li2016,Li2016c,Li2017,Serban2017,Zhang2018b,Olabiyi2018b,Olabiyi2019,Olabiyi2019b}.

In this paper, we propose DLGNet, a transformer-based model for multi-turn dialogue modeling that addresses some of the highlighted problems above. The use of a transformer architecture allows DLGNet to capture long-term temporal dependencies in the dialogue data better than the existing RNN-based architectures \cite{Vaswani2017}. However, applying a vanilla Seq2Seq transformer \cite{Vaswani2017} for dialogue modeling does not work well because of the semantic redundancy in dialogue data. To overcome this, DLGNet models the joint distribution of the context and response instead of the conditional distribution of the response given the context, usually employed in Seq2Seq frameworks \cite{Vinyals2015,Serban2016,Olabiyi2018,Vaswani2017}.  DLGNet also addresses the syntactic redundancy in dialogue data by appending random paddings before and after the input data. This helps to break down the learning barrier from the concave entropy profile of human conversation data, as shown in  Fig. \ref{entropy}. The flattening of the entropy profile also provides regularization during training, and reduces even the extent of the exposure bias problem. Finally, to avoid the out-of-vocabulary problem, DLGNet uses byte pair encoding (BPE) similar to GPT-2 \cite{Radford2019} to provide 100\% coverage for any Unicode input and output texts. 
Given all these proposed changes, we train DLGNet models using only the maximum likelihood objective. 
DLGNet models, despite being trained with only the maximum likelihood objective, demonstrate state-of-the-art performance on the Movie and Ubuntu datasets, as measured in terms of BLEU, ROUGE, and distinct n-gram scores.


\begin{table} 
\caption{Example of generated responses from DLGNet models}
\label{tb:samples_gpt}
\vspace{-10pt}
\begin{center}
\begin{tiny}
\setlength\tabcolsep{2.0pt}
\begin{tabularx}{\linewidth}{ lX }
\toprule
Item & Utterance \\
\midrule
\textbf{Movie}\\
Context\_0 & says he wanted food . $<$person$>$ . he wanted the gold .\\
Context\_1 & how ' s he going to want the gold ? he couldn ' t even know we had it .\\
Groundtruth & he . he could have been watching us . \\
DLGNet-117M \protect\footnotemark & he knew we did . \\
DLGNet-345M & $<$person$>$ ' s just say he doesn ' t have enough to buy a trailer and seeds . i ' ll write him up as a decorated veteran of the united states army . \\
DLGNet-117M\_Fresh \protect\footnotemark & $<$person$>$ , you know the break-in , i might not be able to afford it . \\
DLGNet-345M\_Fresh & are you going to preach now ? \\
\midrule
Context\_0 & what are you thinking ?\\
Context\_1 & that i want you to take me home . \\
Groundtruth & now ? we just -- \\
DLGNet-117M & now ? we ' re going now ? what about tomorrow ? \\
DLGNet-345M & now ? we ' re on the run ! \\
DLGNet-117M\_Fresh & now ? \\
DLGNet-345M\_Fresh & no . keep me company . $<$person$>$ offers its rewards . $<$person$>$ appetit . \\
\toprule
\textbf{Ubuntu}\\
Context\_0 & The netboot one is suppose to download packages from the net. \\
Context\_1 & like the ones to be installed? or the installed to be run? \\
Groundtruth & Installed. The netbook also features the non-graphical installer. \\
DLGNet-117M & the installed to be run. \\
DLGNet-345M& the ones to be installed. \\
DLGNet-117M\_Fresh& I think the netboot one is the one that is installed to the net. \\
DLGNet-345M\_Fresh& the ones to be installed to. \\
\midrule
Context\_0 & DJones: update manager won't detect 12.04.1 as a new version if you are already running 12.04, because 12.04.1 = 12.04 + lots of package updates \\
Groundtruth & DJones I am running 10.04. \\
DLGNet-117M & I'm not sure what you mean by "update manager won't detect" \\
DLGNet-345M & I see. I guess I'll just do a fresh install. \\
DLGNet-117M\_Fresh & I'm running 12.04.1, and I'm not running 12.04.1. \\
DLGNet-345M\_Fresh & I'm not sure what you mean by "update manager" \\
\bottomrule
\end{tabularx}
\vspace{-20pt}
\end{tiny}
\end{center}
\end{table}

\section{Task Description}
\label{gpt2}

Consider a dialogue sample consisting of a sequence of $N$ utterances,
$\mathbf{x}=\big(x_1,x_2,\cdots,x_N\big)$, where each utterance
$x_i=\big(x_i^1,x_i^2,\cdots,x_i^{M_i }\big)$ contains a
variable-length sequence of $M_i$ word tokens such that ${x_i}^j \in V $ for
vocabulary $V$. At any time step $i$, the dialogue history is given
by $\mathbf{x}_i=\big(x_1,x_2,\cdots,x_i\big)$.
The dialogue response generation task can be defined as follows: 
Given a dialogue history $\mathbf{x}_i$, generate a response
$y_i=\big(y_i^1,y_i^2,\cdots,y_i^{T_i}\big)$, where $T_i$ is the
number of generated tokens such that the distribution of the generated
response $P(y_i)$ is indistinguishable from that of the ground truth
$P(x_{i+1})$ and $T_i=M_{i+1}$. 
The distribution of the model output sequence can be factored by the product rule:
\begin{align}
P(y_i|\mathbf{x}_i) = \prod_{j=2}^{T_i}P\big(y_i^j | y_i^{1:j-1}, \mathbf{x}_i\big)
\label{eq:pf}
\end{align}
where $y_i^{1:j-1} = (y_i^1,\cdots,y_i^{j-1})$.

The MLE objective based on the conditional distribution of (\ref{eq:pf}) can be expressed as 
\begin{align}
L_{Cond} = -logP_{\theta}(y_i|\mathbf{x}_i) = -\sum_{j=2}^{T_i}logP_{\theta}\big(y_i^j | y_i^{1:j-1}, \mathbf{x}_i\big)
 \label{eq:mle}
\end{align}
where $\theta$ are the model parameters. 



\addtocounter{footnote}{-1}
\footnotetext{Model with pretraining}
\addtocounter{footnote}{1}
\footnotetext{Model without pretraining}

This formulation, known as Seq2Seq, originated from machine translation \cite{Sutskever2014} and assumes that the context-response pair in the training examples are fairly unique. Seq2Seq is the basis of most of the previous work on dialogue modeling. The framework, however, does not account for the semantic and syntactic redundancy in human conversations as pointed out by \citeauthor{Li2016} \shortcite{Li2016}. 

\section{DLGNet Model Description}
In order to address the semantic redundancy, we propose to jointly model both the context and the response as an alternative to the mutual information objective \cite{Li2016,Zhang2018b}. The resulting distribution and the objective function can then be respectively expressed as
\begin{align}
P(y_i,\mathbf{x}_i) = P(y_i|\mathbf{x}_i)P(\mathbf{x}_i)
\label{eq:DLGNet}
\end{align}

\begin{align}
L_{Joint} = -logP_{\theta}(y_i|\mathbf{x}_i) -logP_{\theta}(\mathbf{x}_i)
 \label{eq:mle}
\end{align}

While (\ref{eq:DLGNet}) addresses the semantic redundancy, it does not address the syntactic redundancy coming from the concave positional entropy profile of dialogue data. To circumvent this, we append random informative paddings (sampled from the dataset) before ($\mathbf{x}_i^b$) and after ($\mathbf{x}_i^a$), the dialogue example of interest, leading to
\begin{align}
P(\mathbf{x}_i^a,y_i,\mathbf{x}_i, \mathbf{x}_i^b) = P(\mathbf{x}_i^a)P(y_i|\mathbf{x}_i)P(\mathbf{x}_i)P(\mathbf{x}_i^b)
\label{eq:DLGNet2}
\end{align}
and 
\begin{align}
L_{DLGNet} = & -logP_{\theta}(\mathbf{x}_i^a) -logP_{\theta}(y_i|\mathbf{x}_i) \nonumber \\ 
& -logP_{\theta}(\mathbf{x}_i) -logP_{\theta}(\mathbf{x}_i^b)
 \label{eq:mle}
\end{align}
since $\mathbf{x}_i^b$ and $\mathbf{x}_i^a$ are independent of $(y_i,\mathbf{x}_i)$. As we see from the resulting entropy profile in Fig. \ref{entropy}, appending random paddings circumvents the adverse effect of syntactic redundancy in dialogue data on model training. The conditional distribution $P(y_i|\mathbf{x}_i)$ in (\ref{eq:pf}) is then just an inference on the joint distribution of (\ref{eq:DLGNet2}).

DLGNet adopts GPT-2's autoregressive transformer architecture \cite{Radford2019} using only the decoder part of the original transformer architecture \cite{Vaswani2017} since there is no need for a separate encoder network.
Autoregressive transformer models use multiple layers of masked multi-head self-attention to map a sequence of input tokens to a sequence of output tokens (i.e., the input sequence token shifted one position to the right). 
During inference, at each step, the model is autoregressive, consuming the previously generated token as additional input when generating the next.
There are some basic conceptual differences between autoregressive architectures based on transformers and those based on recurrent neural networks (RNNs). 
For instance, while the output of an RNN layer depends on only the immediate previous output, a transformer layer output consists of attention over all previous outputs. 
Due to this lack of ordering in transformer architectures, the position representation is usually passed along with the input tokens into the model \cite{Vaswani2017}. 

In order to take advantage and evaluate the impact of pretrained parameters, we use two model configurations 
i.e., 
(i) \textbf{DLGNet-117M} - with 117M parameters, 12 attention layers, and a hidden state size of 767, and 
(ii) \textbf{DLGNet-345M} - with 345M parameters, 24 attention layers, and a hidden state size of 1024; similar to the publicly available GPT-2 models \cite{Radford2019}. 

\begin{table*}[ht]
\caption{Automatic Evaluation of Model Performance}
\label{tb:auto}
\begin{center}
\begin{small}
\vspace{-10pt}
\setlength\tabcolsep{4.0pt}
\begin{tabular}{l|cc|cc||cc|cc}
\toprule
\multirow{3}{*}{\textbf{Model}}  &   \multicolumn{4}{c||}{\textbf{Movie}} & \multicolumn{4}{c}{\textbf{Ubuntu}} \\
& \multicolumn{2}{c|}{Relevance} & \multicolumn{2}{c||}{Diversity} & \multicolumn{2}{c|}{Relevance} & \multicolumn{2}{c}{Diversity} \\    
 & BLEU & ROUGE & DIST-1/2 & NASL & BLEU & ROUGE & DIST-1/2 & NASL \\
\midrule
HRED      & 0.0474   & 0.0384 & 0.0026/0.0056 & 0.535 & 0.0177  & 0.0483  & 0.0203/0.0466 & 0.892 \\
VHRED     & 0.0606   & 0.1181 & 0.0048/0.0163 & 0.831 & 0.0171  & 0.0855  & 0.0297/0.0890 & 0.873 \\
hredGAN\_u & 0.0493   & 0.2416 & 0.0167/0.1306 & 0.884 & 0.0137  & 0.0716  & 0.0260/0.0847 & 1.379 \\
hredGAN\_w & 0.0613 & 0.3244 & 0.0179/0.1720 & \textbf{1.540} & 0.0216  & 0.1168  & 0.0516/0.1821 & 1.098 \\
DAIM & 0.0155 & 0.0077 & 0.0005/0.0006 & 0.721 & 0.0015 & 0.0131 & 0.0013/0.0048 & \textbf{1.626} \\
aBoots\_u\_cat & 0.0880  & 0.4063 & 0.0624/0.3417 & 0.918 & 0.0210   & 0.1491 & 0.0523/0.1795 & 1.040 \\
aBoots\_w\_cat & 0.0940   & 0.3973 & 0.0613/0.3476 & 1.016 & 0.0233 & 0.2292 & 0.1288/0.5190 & 1.208 \\
\midrule
\midrule

DLGNet-117M\_Fresh & 0.1796   & 0.4338 & 0.1198/0.4578 & 1.011 & 0.0215 & 0.1978 & 0.1827/0.4074 & 0.829 \\
DLGNet-345M\_Fresh & 0.2682   & 0.4881 & 0.1286/0.4612 & 0.907 & \textbf{0.0315} & 0.2041 & 0.1927/0.4468 & 0.794 \\
\midrule

\textbf{DLGNet-117M} & 0.1872   & 0.4346 & 0.1232/0.4506 & 0.982 & 0.0279 & 0.2191 & 0.2228/0.4953 & 0.746 \\
\textbf{DLGNet-345M} & \textbf{0.2742}   & \textbf{0.4945} & \textbf{0.1282/0.4736} & 0.895 & \textbf{0.0309} & \textbf{0.2409} & \textbf{0.2436/0.5632} & 0.759 \\

\bottomrule
\end{tabular}
\end{small}
\end{center}
\vspace{-10pt}
\end{table*}

\section{Model Training}
We trained the small DLGNet-117M and the medium DLGNet-345M models on multi-turn dialogue datasets initialized with either random noise or pretrained language model parameters. The models are trained end-to-end 
using the Adaptive Moment Estimation (Adam) stochastic gradient descent algorithm with a learning rate of 0.001. The maximum sequence length is 1024. Due to GPU memory limitations, we use a batch size of 2 and accumulate 
gradients over 5 iterations, making the effective batch size 10. Both models are trained until the training perplexity on the dialogue datasets reaches a steady state. Finally, the models are implemented, trained, and evaluated using Python and the TensorFlow deep learning framework.

\section{Experiments}
\label{experiments}

\subsection{Setup}
We evaluated DLGNet models on the Movie Triples and Ubuntu Dialogue corpora randomly split into training, validation, and test sets, using 90\%, 5\%, and 5\% proportions. Since we use BPE with 100\% tokenization coverage, we performed no preprocessing of the datasets whatsoever. For each training example, however, we randomly sample a target conversation and two padding chunks from the dataset to fill up the maximum input sequence length. We append the paddings to the target conversation, one before, and one after, separated by a unique token. The target conversation in each training example in turn consists of utterances that are separated by another unique token.

The Movie dataset \cite{Serban2016} spans a wide range of topics with few spelling mistakes and contains about 240,000 dialogue triples, which makes it suitable for studying the relevance-diversity tradeoff in multi-turn conversations \cite{Zhang2018b}. The Ubuntu dialog dataset extracted from the Ubuntu 
Relay Chat Channel \cite{Serban2017} contains about 1.85 million conversations with an average of 5 utterances per conversation. This dataset is ideal for training dialogue models that can provide expert knowledge/recommendation in domain-specific conversations.

We compare DLGNet multi-turn dialogue performance with existing state-of-the-art dialogue models including (V)HRED\footnote{implementation obtained from \url{https://github.com/julianser/hed-dlg-truncated}} \cite{Serban2016,Serban2017}, DAIM\footnote{implementation obtained from \url{https://github.com/dreasysnail/converse_GAN}} \cite{Zhang2018b}, hredGAN \cite{Olabiyi2018}, and aBoots \cite{Olabiyi2019b}. Note that DAIM is single turn and does not use a multi-turn dialogue context, but we have included it here for completeness. We compare how the models perform based on informativeness (a combination of relevance and diversity metrics) of generated responses. For relevance, we adopted BLEU-2 \cite{Papineni2002} and ROUGE-2 \cite{Lin2014} scores. For diversity, we adopted distinct unigram (DIST-1) and bigram (DIST-2) \cite{Li2016} scores as well as normalized average sequence length (NASL), similar to \citeauthor{Olabiyi2018} \shortcite{Olabiyi2018}.


All models are evaluated in autoregressive mode, i.e., we pass a multi-turn dialogue context to the model inputs and the models generate a sequence of response tokens using the context and all the previously generated tokens until the end-of-sequence token is reached. All models are greedily sampled to generate the model outputs, with the exception of the aBoots and DLGNet models. In these cases, we first search for the optimum top\_k between 0 and 20 inclusive that maximizes
the overall BLEU-2 (relevance) score of the validation set using the top\_k sampling strategy \cite{Radford2019}. It turns out that for all DLGNet models, the optimum top\_k is 1 across datasets, which is equivalent to greedy sampling. For aBoots, however, we noted different
optimum values for different model configurations for the Ubuntu dataset. This may be due to the effect of model bootstrapping on the model output distribution, especially on a large and domain-specific dataset like Ubuntu. 
For detailed information, we depict the trajectory of the hyper-parameter selection metric with increasing top\_k values in Figure \ref{topk_selection}.

\section{Results and Discussion}
\label{res_dis}

\subsection{Quantitative Evaluation}
\label{eval_quant}
We report the quantitative measures in Table \ref{tb:auto}.  The transformer-based DLGNet provides a significant improvement in response generation performance over existing methods such as (V)HRED, hredGAN, DAIM, and adversarial bootstrapping (aBoots), all of which are based on recurrent neural networks. In fact, DLGNet achieves the best performance to date on the Movie triples and Ubuntu dialogue datasets in terms of BLEU, ROUGE, and distinct n-gram scores. This indicates that, despite being trained only with the maximum likelihood objective, the autoregressive transformer architecture in conjunction with the random padding injection, is able to overcome some of the problems that have plagued existing dialogue models such as semantic and syntactic redundancy, and exposure bias. Also contributing to the models' performance improvement is the 100\% input coverage from the BPE encoding, which eliminates the generation of $<$UNK$>$ tokens (this is especially helpful for the Ubuntu dataset with a large number of out-of-vocabulary tokens) as well as the joint modeling of the context and response. Also, in contrast to existing work reporting a trade-off between relevance and diversity \cite{Zhang2018b,Li2016,Li2016c}, we observe that relevance performance improves with diversity performance in DLGNet models. It is worth pointing out, however, that DLGNet models tend to generate shorter responses than adversarially trained models (hredGAN and aBoots). This indicates that the models still suffer from the impact of using only the maximum likelihood training objective. Alleviating this problem with an adversarial training objective similar to aBoots and or hredGAN should further improve performance and will be considered in our future work.            

\subsection{Qualitative Evaluation}
\label{eval_quanl}
Random samples of the model outputs are shown in Tables \ref{tb:samples_gpt} and \ref{tb:samples_all}. One striking observation is the high level of coherence in the generated responses from DLGNet models. The models are able to capture both short- and long-term temporal dependencies in their responses. The models give responses that are relevant to the topic of the discussion, and are able to answer posed questions with answer choices. Also, they don't simply generate the all-too-common phrase ``I'm not sure'' like existing models; they are able to point to areas of the context they are uncertain about (see the Ubuntu section of Table \ref{tb:samples_gpt}). 


\begin{table*}[ht]
\caption{Ablation Performance of DLGNet Models with Static Padding} 
\label{tb:ablation}
\begin{center}
\begin{small}
\vspace{-10pt}
\setlength\tabcolsep{4.0pt}
\begin{tabular}{l|cc|cc||cc|cc}
\toprule
\multirow{3}{*}{\textbf{Model}}  &   \multicolumn{4}{c||}{\textbf{Movie}} & \multicolumn{4}{c}{\textbf{Ubuntu}} \\
& \multicolumn{2}{c|}{Relevance} & \multicolumn{2}{c||}{Diversity} & \multicolumn{2}{c|}{Relevance} & \multicolumn{2}{c}{Diversity} \\    
 & BLEU & ROUGE & DIST-1/2 & NASL & BLEU & ROUGE & DIST-1/2 & NASL \\
\midrule
\textbf{DLGNet-117M} &&&&&&&&\\
Single-turn Joint with BPE      & $\sim$0.0   & $\sim$0.0 & 0.0400/0.1502 & 0.072 & $\sim$0.0  & 0.0004  & 0.1946/0.4636 & 0.064 \\
Single-turn Conditional with BPE & 0.0013   & 0.0296 & 0.0134/0.0482 & 3.582 & $\sim$0.0  & 0.0083  & 0.0723/0.1470 & 0.890 \\
Multi-turn Joint with BPE     & 0.1825   & 0.1321 & 0.0346/0.0838 & 0.610 & 0.0012  & 0.1172  & 0.1719/0.3482 & 0.2937 \\
Multi-turn Conditional with BPE & 0.0096 & 0.0628 & 0.0088/0.0394 & 3.425 & 0.0048  & 0.0766  & 0.0500/0.1454 & 2.372 \\
Multi-turn Joint with Basic Tokenizer     & 0.0518   & 0.0630 & 0.0176/0.0540 & 1.101 & 0.0030  & 0.0384  & 0.0465/0.0949 & 0.566 \\

Multi-turn Conditional with Basic Tokenizer & 0.0149 & 0.1628 & 0.0394/0.1770 & 1.472 & $\sim$0.0  & 0.0136  & 0.2211/0.4192 & 0.281 \\
\midrule
\midrule
\textbf{DLGNet-345M} &&&&&&&&\\
Single-turn Joint with BPE     & $\sim$0.0   & $\sim$0.0 & $\sim$0.0/$\sim$0.0 & 0.072 & $\sim$0.0  & 0.0006  & 0.4741/0.9760 & 0.061 \\
Single-turn Conditional with BPE & 0.0006   & 0.0212 & 0.0010/0.0419 & 3.582 & 0.0004  & 0.0158  & 0.0721/0.1671 & 3.437 \\
Multi-turn Joint with BPE    & 0.0449   & 0.1931 & 0.0460/0.1273 & 0.531 & $\sim$0.0  & 0.0121  & 0.3323/0.4406 & 0.227 \\
Multi-turn Conditional with BPE & 0.0010 & 0.0125 & 0.0091/0.0422 & 3.918 & 0.0004  & 0.0158  & 0.0721/0.1671 & 4.108 \\

Multi-turn Joint with Basic Tokenizer     & 0.0376   & 0.1389 & 0.0232/0.0654 & 0.543 & 0.0042  & 0.0341  & 0.0568/0.1299 & 0.552 \\
Multi-turn Conditional with Basic Tokenizer & 0.0057 & 0.0970 & 0.1568/0.3785 & 0.331 & 0.0015  & 0.0345  & 0.1555/0.3990 & 0.470 \\

\bottomrule
\end{tabular}
\end{small}
\end{center}
\vspace{-10pt}
\end{table*}

\section{Ablation Studies on DLGNet Models with Random Informative Padding}
\label{ablation}
In this section, we carry out a more detailed analysis and discussion of different configurations of DLGNet models as well as their performance across datasets, using the evaluation results in Table \ref{tb:auto}.

\subsection{Open vs. Closed Domain Dataset}
From Table \ref{tb:auto}, we observe that the performance improvement achieved by DLGNet models over existing models is higher for the open-domain Movie Triples dataset than for the closed-domain Ubuntu Dialogue dataset with or without pretraining. While the performance difference could be due to the size of the dataset, it could also indicate that closed-domain dialogue responses are inherently more difficult to learn, even for large and expressive models such as the DLGNet transformer.

\subsection{Effect of Model Pretraining}
Although models with pretraining generally perform better than ones trained with random initialization, we observe that the performance difference is not significant. This shows that the performance of the DLGNet is mostly due to the multi-layer self attention model architecture rather than the scaffolding achieved from language model pretraining. We observe similar behavior across datasets. However, pretraining seems to be consistently more helpful for open-domain datasets versus closed-domain datasets. This might be because the distribution of the language data used for pretraining is similar to the open-domain dataset but different from the closed-domain dataset. Also, models without pretraining tend to generate longer responses on average compare to those with pretraining. This indicates that model pretraining also plays a role in the relevance-diversity tradeoff.  

\subsection{Effect of Model Size}
We also compare the small (DLGNet-117M) and large (DLGNet-345M) models. We observe that there is a significant performance improvement of the larger over the smaller model on the Movie dataset (about 50\%), but a smaller performance improvement on the Ubuntu dataset. It's also surprising that the larger model doesn't overfit to the Movie dataset. Overfitting might have been prevented by the injection of random padding into the input data, which regularizes the model training by artificially inducing high entropy into the data.

\subsection{Relevance vs. Diversity Tradeoff}
\begin{figure}[ht]
\begin{center}
\centerline{\includegraphics[width=\columnwidth]{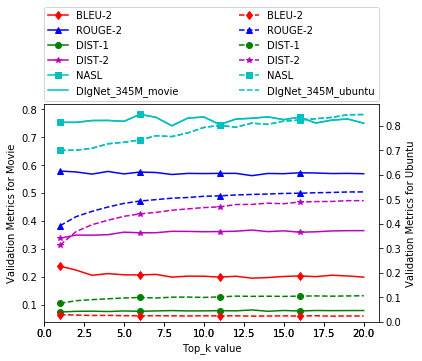}}
\caption{Relevance vs. diversity tradeoff with top\_k sampling for DLGNet-345M models.}
\label{rel_div_top_k}
\end{center}
\vskip -0.2in
\end{figure}

\begin{figure}[ht]
\begin{center}
\centerline{\includegraphics[width=\columnwidth]{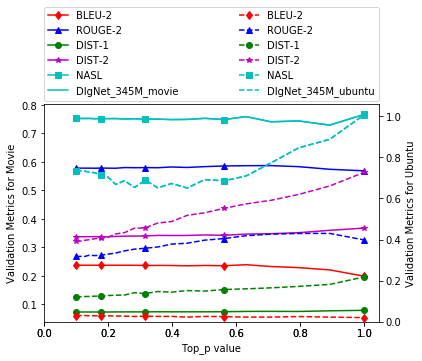}}
\caption{Relevance vs. diversity tradeoff with top\_p sampling for DLGNet-345M models.}
\label{rel_div_top_p}
\end{center}
\vskip -0.2in
\end{figure}

The results in Table \ref{tb:auto} show state-of-the-art relevance performance with some compromise on the response length. Here, we explore the possibility of generating longer and more diverse responses with the trained models and estimate the effect on the relevance scores. For this experiment, we chose the larger DLGNet-345M models of both datasets and tried two sampling techniques, i.e., top\_k \cite{Radford2019} and top\_p nucleus \cite{Holtzman2019,Zellers2019} sampling strategies on the validation sets. 
The trajectory of the evaluation metrics with increasing top\_k and top\_p values are shown Figs. \ref{rel_div_top_k} and \ref{rel_div_top_p} respectively.
With top\_k sampling, increasing the top\_k value increases the response length at the expense of relevance metrics like BLEU for both datasets, as expected. However, the response length increase is more significant on the Ubuntu dataset than the Movie dataset. It is also surprising that the ROGUE-2 score for Ubuntu increases with increasing top\_k value, which is the reverse of the case for the Movie dataset. Also, Fig. \ref{rel_div_top_k} shows that it is more advantageous to trade off relevance for diversity on the Ubuntu dataset compare to the Movie dataset. This is probably due to the size and closed-domain nature of the Ubuntu dataset, which makes it more difficult to learn with the maximum likelihood estimation only.

We observe a similar pattern with the top\_p nucleus sampling in Fig. \ref{rel_div_top_p}. This reinforces the fact that greedy sampling may be sufficient for open-domain datasets such as Movie.   

\section{Further Ablation Studies on DLGNet Models}
We also set out to analyze the features of DLGNet that make it suitable for multi-turn dialogue modeling.
We train both DLGNet-117M and DLGNet-345M models on both datasets, but replace the random informative paddings with static paddings using a pad token. Below are the definitions of the model configuration factors considered:

1.) Multi-turn Data: Training data is variable-length multi-turn data padded to a fixed length. This helps to evaluate the effect of using random informative padding.

2.) Single-turn Data: Training data is variable-length single-turn data padded to a fixed length. This helps to evaluate the effect of number of turns.

3.) Joint model: DLGNet models are trained by jointly modeling the dialogue context and response.

4.) Conditional model: DLGNet models are trained in the traditional sequence-to-sequence mode with a bidirectional encoder and an autoregressive decoder for a conditional modeling of the dialogue response given the context \cite{Vaswani2017}. 

5.) Basic Tokenizer: We use a basic tokenization traditionally used in dialogue modeling instead of BPE tokenization to evaluate the effect of tokenization coverage. It also provides an apples-to-apples comparison between the transformer-based and RNN-based architectures.

\subsection{Effect of Random Padding Injection}
The results in Table \ref{tb:ablation} are from models trained with static paddings. The models perform significantly worse than those of Table \ref{tb:auto}. Without random padding injection, the models quickly overfit to the low entropy regions of the training data, which leads generic and/or short responses.

\subsection{Single Turn vs. Multi-turn}
We also observe that the multi-turn models perform better than single-turn models on BPE tokenized data. This is expected because the multi-turn models capture longer temporal dependencies in the input data. It is also worth mentioning that the single-turn performance is further hurt by BPE tokenization since it tends to work better with long input sequences.

\subsection{Joint vs. Conditional Models}
For multi-turn models, the joint modeling architecture yields better performance than the conditional Seq2Seq architecture. This trend is however reversed for single-turn models. This is because a model that focuses on jointly modeling both the context and the response performs better with longer contextual information compared to a model that focuses on modeling only the conditional distribution of the response given the context. Therefore, multi-turn dialogue model should rather employ the joint structure instead of the conditional Seq2Seq structure. 

\subsection{Effect of Tokenization Coverage}
For a more fair comparison with previous work on multi-turn dialogue not using random padding injection and 100\% BPE tokenization, we trained the DLGNet models on multi-turn data with basic tokenization. The tokenization coverages of the basic tokenizer used are 83.9\% and 4.19\% for Movie and Ubuntu datasets respectively. Basically, most of the Ubuntu tokens are mapped to the $<$UNK$>$ token. In comparison with previous work on HRED, the results in Table \ref{tb:ablation} show that the transformer-based DLGNet models under the same conditions perform better than the basic HRED model but worse than the improved HRED models (such as VHRED, hredGAN, and aBoots). In comparison with other transformer-based configurations, the smaller size multi-turn models perform better than their BPE counterparts but the larger size models perform worse. This is probably due to the overfitting of the larger models.

\section{Conclusion}
\label{concl}
In this paper, we have proposed DLGNet, an extension of autoregressive transformer models such as GPT-2 for multi-turn dialogue modeling. Our experiments show that DLGNet models perform better than existing state-of-the-art multi-turn dialogue models. They also achieve the best performance to date on open-domain Movie and closed-domain Ubuntu datasets based on BLEU, ROUGE and distinct n-gram scores. Our experiments reveal that the combination of (i) the transformer architecture with (ii) the injection of random paddings exploiting the large maximum input sequence is responsible for the performance improvement over existing methods. Other contributing factors include joint modeling of dialogue context and response, and the 100\% tokenization coverage from the byte pair encoding (BPE). Our analysis also reveals some tradeoffs between response relevance and response length, and we showed how different sampling strategies can be used to make an informed decision about such response relevance-diversity compromises.  In our future work, we plan to investigate how to improve on the length of the generated responses without necessarily sacrificing their coherence and their relevance to the dialogue context. 

\iftrue
\begin{table}[t]
\caption{ Sample responses of DAIM, (V)HRED, hredGAN, aBoots and DLGNet models.}
\label{tb:samples_all}
\begin{center}
\begin{tiny}
\setlength\tabcolsep{6.0pt} 
\begin{tabularx}{\linewidth}{ l X }
\toprule
\textbf{Model} & \textbf{Response} \\
\textbf{Movie} \\
\toprule
\textbf{Context\_0}& we ' re grateful , $<$person$>$ . it ' s not that . those craftsman tools . they ' re a dream come true .\\
\textbf{Context\_1}& your dad and i were raised with a certain work ethic , that ' s all .\\
\textbf{Groundtruth} & we can ' t start living like we ' re the rockerfellers or something . \\
HRED & $<$person$>$ .\\
VHRED & $<$person$>$ , mr . $<$person$>$ . $<$person$>$ , i ' m not going to tell you , $<$person$>$ . i ' m sorry . i ' m sorry , $<$person$>$ . i ' m sorry \\
hredGAN\_w & $<$person$>$ , i know you ' re see . $<$ continued\_utterance $>$ $<$person$>$ , please .\\
aBoots\_w\_cat & you ' re a liar ! \\
DLGNet-117M & $<$person$>$ , come on , hang on . how do you like that ? i can barely believe you ' re doing this . \\
DLGNet-345M & i have some very bad dreams . \\
DLGNet-117M\_Fresh & i ' m sorry about all this . $<$person$>$ , this is all their house . \\
DLGNet-345M\_Fresh & remember that shoe you lost . \\
\midrule 
\textbf{Context\_0}& $<$person$>$ i ' ve raised my rate . i might even cut my sessions to forty minutes .\\
\textbf{Context\_1}& $<$person$>$ , i know you don ' t like flattery but if anybody i know is going to win a nobel prize , it ' s you . you got ta be thinking about your next book so i know you ' ll find this case particularly interesting . \\
\textbf{Groundtruth} & what sort of case is it , $<$person$>$ ? \\ 
HRED&  i ' m sorry , $<p$erson$>$ . \\
VHRED:& it ' s not a lot of $<$unk$>$ . i ' m not going to tell you , $<$person$>$ . i ' m not going to help you . \\
hredGAN\_w& i ' m a little nervous , but i don ' t care if there ' s something more important than $<$person$>$ . \\
aBoots\_w\_cat& if i ' m sure they ' ll call it ' s place $<$person$>$ . \\
DLGNet-117M & what sort of case is it , $<$person$>$ ? \\
DLGNet-345M & what sort of case is it , $<$person$>$ ? \\
DLGNet-117M\_Fresh & what sort of case is your next ? \\
DLGNet-345M\_Fresh & like what ? \\
\midrule
\textbf{Context\_0}& what was that for ?\\
\textbf{Context\_1}& i ' d forgotten how exciting it is working together .\\
\textbf{Groundtruth} & you ' ve been so quiet lately , $<$person$>$ , i thought you no longer cared . \\
HRED& i ' m sorry , $<$person$>$ . \\
VHRED:& $<$person$>$ , $<$person$>$ . i don ' t know .\\
hredGAN\_w & i ' m sorry , $<$person$>$ . so many people are so interested , but you ' re a very nice guy .\\
aBoots\_w\_cat& we ' re going to take a look ?\\
DLGNet-117M & you look like working with $<$person$>$ . \\
DLGNet-345M & you ' re a helluva translator . \\
DLGNet-117M\_Fresh & you ' ve been so nice to $<$person$>$ ach . \\
DLGNet-345M\_Fresh & oh .\\
\toprule
\textbf{Ubuntu} \\
\toprule
\textbf{Context\_0}& DJones: update manager won't detect 12.04.1 as a new version if you are already running 12.04, because 12.04.1 = 12.04 + lots of package updates\\
\textbf{Context\_1}& DJones I am running 10.04.\\
\textbf{Context\_2}& what does 'gksu update-manager \-c' report?\\
\textbf{Groundtruth} & does not report that there is a new version out. So I am just holding. No rush. \\
HRED &  I don't know what the problem is \\
VHRED &  you mean ? \\
hredGAN\_w & yes , I get the new one , I just want to get the new one \\
aBoots\_w\_cat & it says : no such file or directory , that is what it says \\
DLGNet-117M & gksu: command not found \\
DLGNet-345M & it says No new release found \\
DLGNet-117M\_Fresh & http://pastebin.com/m7e8e0e7c \\
DLGNet-345M\_Fresh & it says "update-manager: error: no new version is available" \\
\midrule
\textbf{Context\_0}& libgnome2-dev is the package that has that pkgconfig file\\
\textbf{Context\_1}& After I installed libgnome2-dev I got a different output. http://pastebin.com/zpDeGs81 sorry it is so long.\\
\textbf{Context\_2}& I thought that was what was going on thanks. :) I ran the command ./configure \&\& sudo make \&\& sudo make install would that have worked.\\
\textbf{Groundtruth} & you shouldn't run: sudo make; but yes \\
HRED &   I don't know what the problem is \\
VHRED &  what do you mean ? \_\_eou\_\_ what are you trying ? \\
hredGAN\_w & I have no idea , I can just copy the files and copy files to the other one \\
aBoots\_w\_cat & yeah , if you have a few times to put things in the ' make ' command it is not in a file you can install it by hand . \\
DLGNet-117M & no, that's not what you want to do \\
DLGNet-345M & no, you need to run it as root \\
DLGNet-117M\_Fresh & no, it won't. \\
DLGNet-345M\_Fresh & yes, that's what I did \\
\bottomrule
\end{tabularx}
\end{tiny}
\end{center}
\vspace{-5pt}
\end{table}
\fi

\clearpage
\bibliography{aaai_2020_gpt2}
\bibliographystyle{aaai}

\end{document}